%% file: paper.tex
\pdfoutput=1
\documentclass[runningheads]{llncs}
\usepackage{float}
\usepackage{graphicx}
\usepackage{amssymb}
\usepackage{amsmath,tabularx}
\usepackage{xcolor}
\usepackage{multirow}
\usepackage{siunitx}
\usepackage{booktabs}
\usepackage{etoolbox}
\usepackage{color}
\usepackage{lipsum}
\usepackage{indentfirst}

\definecolor{ben}{rgb}{0.9,0.,0.5}

\definecolor{sher}{rgb}{0.0,0.3,0.5}

\definecolor{todo}{rgb}{1.0, 0., 0.}

\newcommand\asteriskfill{\leavevmode\xleaders\hbox{$\ast\ $}\hfill\kern0pt}

\begin{document}
\title{BFS-Net: Weakly Supervised Cell Instance Segmentation from Bright-Field Microscopy Z-Stacks}
%
\titlerunning{BFS-Net. Weakly Supervised Bright-Field Cell Segmentation}
%
\author{\asteriskfill \\ \asteriskfill}
\author{Shervin Dehghani\inst{1} \and
Benjamin Busam \inst{1} \and
Nassir Navab\inst{1, 2} \and
M. Ali Nasseri\inst{1, 3}}

\authorrunning{S. Dehghani et al.}
%
\institute{
\asteriskfill \\ \asteriskfill \\ \asteriskfill
}
\institute{Computer Aided Medical Procedures, Technical University of Munich, Germany \\ \email{\{shervin.dehghani, b.busam, nassir.navab, ali.nasseri\}@tum.de}\and
Computer Aided Medical Procedures, Johns Hopkins
University, Baltimore, USA \and
Augenklinik rechts der Isar, Technical University of Munich, Germany}
\maketitle              
\input{0-0-abstract}

\section{Introduction}
\input{1-0-introduction}

\section{Related Works}
\input{2-0-related-works}


\section{Methodology}
\input{4-0-methodology}

\section{Experiments and Results}
\input{5-0-experiments}

\section{Discussion}
\input{6-0-discussion}

\bibliographystyle{ieeetr}
\bibliography{literature}

\end{document}

%% file: 0-0-abstract.tex
\begin{abstract}
Despite its broad availability, volumetric information acquisition from Bright-Field Microscopy (BFM) is inherently difficult due to the projective nature of the acquisition process.
We investigate the prediction of 3D cell instances from a set of BFM Z-Stack images.
We propose a novel two-stage weakly supervised method for volumetric instance segmentation of cells which only requires approximate cell centroids annotation.
Created pseudo-labels are thereby refined with a novel refinement loss with Z-stack guidance.
Evaluation shows that our approach can generalize not only to BFM Z-Stack data, but to other 3D cell imaging modalities.
A comparison of our pipeline against fully supervised methods indicates that the significant gain in reduced data collection and labelling results in minor performance difference.

\keywords{Bright-Field Microscopy \and Z-Stack \and Cell Instance Segmentation \and Weak Supervision}

\end{abstract}

%% file: 1-0-introduction.tex
\begin{figure}[htpb]
\centering
\includegraphics[width=\textwidth]{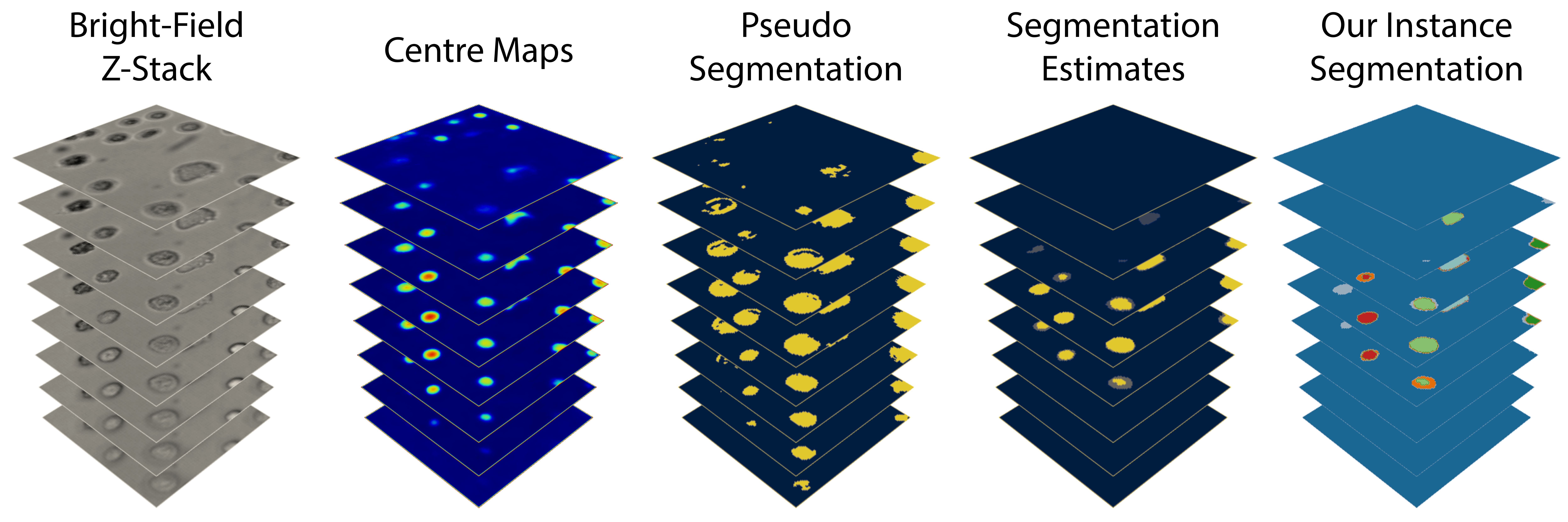}
\caption{\textbf{BFS-Net Data Flow.} We learn from a Bright-Field Z-Stack of cells (1st from left) to predict 3D Centres (2nd) which create Pseudo Segmentation labels (3rd) to learn 3D Segmentation (4th). The estimates are processed to provide 3D Instance Segmentations (right).}
\label{pics:result_overview}
\end{figure}

Automatic micromanipulation of biological cells in 3D space, gained significant interests in the recent years for various applications~\cite{shishkin2020auxiliary}.
The level and precision of automation for micromanipulation tasks is highly dependant on the performance of the 3D localization components in the micromanipulation setup.
Such a localization methods are often utilized to plan the manipulator trajectory, preview the operation and execute the manipulation tasks.
One of the critical aspects for spatial manipulation of cells is reasoning on the distribution of cells at different depths within the situs which is a challenging task from a single 2D microscopy image due to the projective nature of the acquisition process.
Due to the inherent ambiguities of such an estimation, one cannot uniquely determine the 3D location of the cells from a single all-in-focus 2D image~\cite{saxena2007depth}.

One way to address this is by using Electron Microscopy.
While providing highly accurate resolution, such hardware can be costly and is not broadly available.
Moreover, it needs to be used in a specific environmental setup that requires an isolation chamber typically without access from the outside during a scan~\cite{goodhew2000electron}.
In order to be able to manipulate cells reasonably fast and with acceptable performance and speed, an open environment is essential such that a manipulator can be integrated seamlessly.
In practice, Bright-Field Microscopy (BFM) is arguably the most widely spread modality for cellular imaging with its comparably lower cost.
While naturally being available in many biology and medical labs, it also has the potential of volumetric imaging by integration of a micron precision step motor for controlling the Z plane~\cite{lugagne2018identification}.
The captured 3D volume from such an enhanced BFM is different than a depth-aware 3D volume as can be seen in the example in Fig.~\ref{pics:result_overview}.
Due to the effect of BFM-projection in Z direction, the cell image is composed of information from in-focus and out-of-focus parts within the visible area. Different regions are in focus, depending on the Z position within the stack.
While being of projective nature, the data provides focus cues that can be leveraged to estimate depth together with the necessary morphological features to spatially separate the cells~\cite{subbarao1994depth}.

Annotation of instance volumes can be intricate and time-consuming. We therefore leverage BFM Z-Stacks and propose a weakly supervised approach for segmentation of cells within the volume. We first train a cell centre prediction network which we use to create pseudo labels. These labels are used jontly with an image-guided refinement loss to train our BFS-Net segmentation model.
To this end, our main contributions are:
\begin{itemize}
    \item We propose the first weakly supervised instance segmentation method for cells from BFM Z-Stack inputs that only needs cell centre annotations for training.
    \item We propose a novel Z-Stack guided refinement loss which is used to train a 3D segmentation network with centre coordinate induced pseudo labels.
    \item We make the BFM Z-Stack cell dataset publically available to stimulate research in this new domain.
    
\end{itemize}

%% file: 2-0-related-works.tex
\noindent\textbf{Z-Stack Cell Segmentation.} Despite its wide availability surprisingly little work investigates BFM Z-Stacks. The pioneering work of \cite{lugagne2018identification} investigates instance segmentation from this modality. In comparison to our approach, however, they (1) train with full supervision and (2) assign the same label to each pixel coordinate in the entire stack. We believe that the 3D local neighbourhood can be advantageous and estimate volumetric labels.\\ \\
\noindent\textbf{2D/3D Cell Segmentation.} 2D cell instance segmentation methods incorporate variants of U-Net~\cite{ronneberger2015u} combined with post-processing~\cite{al2018deep} steps. The main challenge for these pipelines is to separate the cells and achieve instance-wise precision. 3D cell segmentation approaches similarly use V-Net~\cite{vnet} or 3DUNet~\cite{cciccek20163d} backbones. The recent segmentation work of~\cite{tokuoka20203d} estimates instances in a post post-processing stage taking advantage of each cell center. \\ \\ 
\noindent\textbf{Weakly Supervised Cell Segmentation.} Weakly supervised methods can be separated into two branches based on the used labels being either bounding boxes or cell centroids. \cite{qu2019weakly} uses point annotations as supervision and classifies each pixel based on a combination of k-means and Voronoi tessellation which results in imprecise predictions close to cell edges. \cite{yoo2019pseudoedgenet} attempts to solve this issue trough consecutive edge detection. Another branch of works is based around the idea of \cite{zhou2018weakly}, who introduce Peak Response Maps (PRM). PRMs indicate relevant visual regions created by backpropagation of peak responses from a classifier similar to saliency maps~\cite{selvaraju2017grad}. \cite{nishimura2019weakly} and \cite{dong2019instance} use PRMs with center points and \cite{zhao2018deep} with bounding box annotation to generate segmentation through these visual cues.
This family of methods need a post-processing step to refine the results, and require $\mathcal{O}(n)$ back-propagations at inference which significantly hampers their efficiency for the prediction of $n$ cells in one input volume.
We utilize PRMs only in the first stage of our training to create pseudo labels which are refined in a second stage omitting the costly back-propagation at test time.

%% file: 4-0-methodology.tex
\begin{figure}[htpb]
\centering
\includegraphics[width=\textwidth]{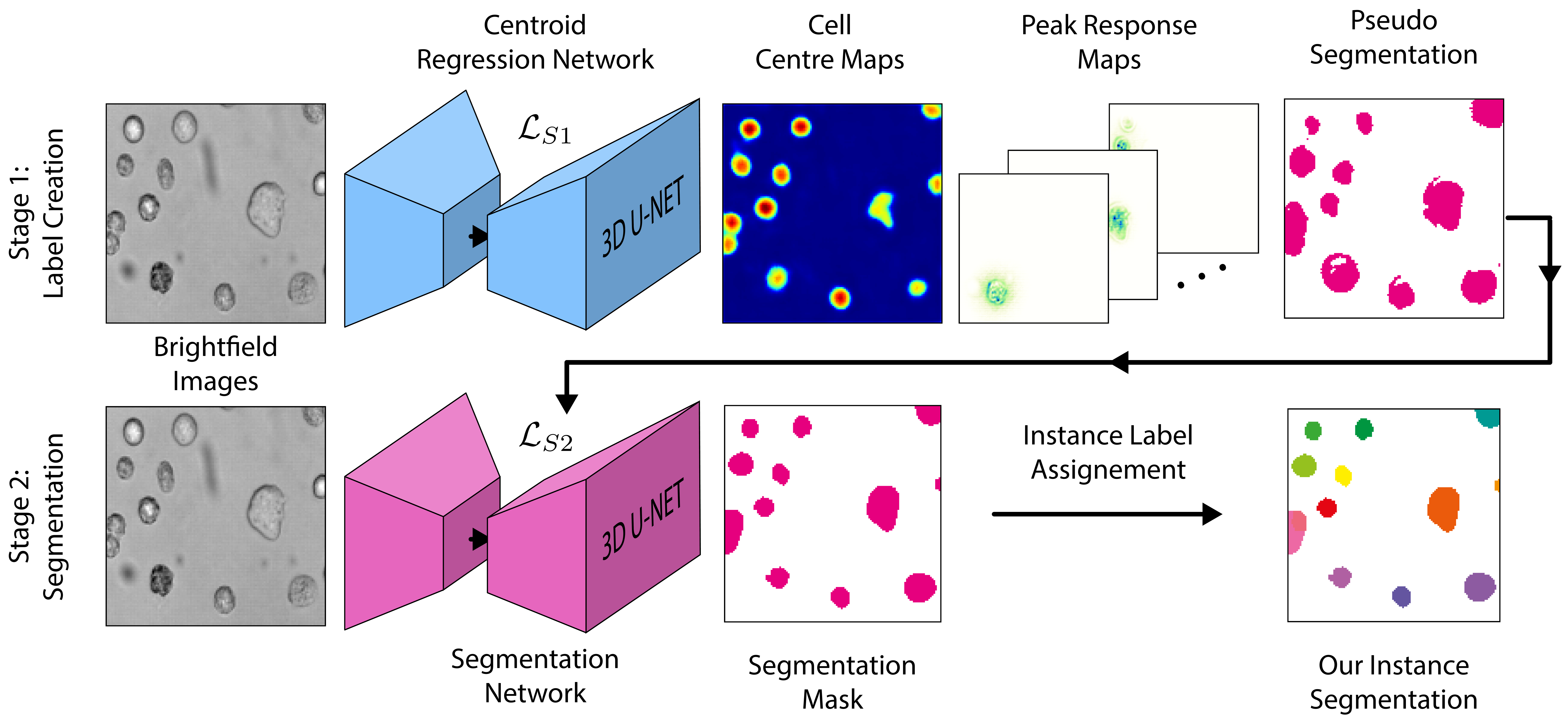}
\caption{\textbf{Pipeline Overview.} Creation of Pseudo Labels from centroid annotations (Row 1), Training with Pseudo Labels to achieve refined segmentation mask and creating the Instance Segmentations (Row 2). For simplification only one slice of the volume is shown.}
\label{pics:brightfield_steps}
\end{figure}
\label{method}
An overview of our pipeline is illustrated in Fig.~\ref{pics:brightfield_steps}. We first train a regression model $S1$ to predict cell center likelihood maps. With $S1$, we generate the Response Map of each of the local maxima of the likelihood map to obtain pseudo segmentation labels.
Previous methods such as \cite{nishimura2019weakly} and \cite{dong2019instance} improve upon back-propagated labels at test time. To omit this additional computational cost and further improve the results, we use a second training stage to train the model $S2$ that learns with these labels and refines the results under Z-Stack image guidance.
With the segmentation mask from $S2$ and the predicted center maps from $S1$, we consecutively apply an efficient label assignment to separate instances.
We detail each step hereafter.

\subsection{Creation of Cell Centre Maps}
Similar to \cite{kainz2015you} we generate a Gaussian based voxel map around each center point. In addition, we divide the space into Voronoi sub-spaces.
The borders of these sub-spaces are hard-negative cases, and we override the Gaussian value with $0$. This helps learning the cell centers in a way which they are separable, even if being adjacent. The voxel likelihood map can be formulated as follows:

\begin{equation}
p(x) = 
\begin{cases}
    0, & \text{if}\ x \in \mathcal{B_V} \\
    e^{-k\frac{\mathcal{D}(x)}{d_m}},& \text{else if}\ \mathcal{D}(x) \leq d_m \\
    0              & \text{otherwise}
\end{cases}
\end{equation}

\noindent where $\mathcal{B_V}$ is the set of voxels which are on Voronoi borders in the volume $\mathcal{V}$, $\mathcal{D}(x)$ is the Euclidean distance of voxel $x$ to the closest center point, and $d_m$ and $k$ are the parameters to control the distribution.

\subsection{Centroid Regression Network and Pseudo Labels}
\label{pseudo}

With the cell centre maps, we train a regression 3DUNet $S1$ for cell center likelihood maps. The training loss for this stage is:
\begin{equation}
    \mathcal{L}_{S1} = \mathcal{L}_{bce} + \lambda\mathcal{L}_{focal}
\end{equation}
where $\mathcal{L}_{bce}$ is a Binary Cross Entropy loss and $\mathcal{L}_{focal}$ is a focal loss, applied on voxels with $p(x) > 0.7$. As in~\cite{zhou2018weakly}, we are interested to have the PRMs of the model output. For each of the local maxima stimulated from model's output, we run a back-propagation through $S1$ to get the visual cue from the input which results in a predicted local peak. Each of these masks belong to a separate instance, but one instance can have multiple masks. Similar to~\cite{dong2019instance}, we use a back-propagation method in 3D which extends the 2D proposal of~\cite{zhao2018deep}. Although being very informative, these response maps cannot be used as a final segmentation result, since they lack sufficient precision. While others use these masks and perform post-processing with e.g. Graph Cuts~\cite{nishimura2019weakly}, we store them as pseudo labels to train a second network $S2$ to directly predict refined peak response maps in one single pass without the need for back-propagation at inference time.

\subsection{Segmentation Network}
\label{segmentation}
At this stage we train a model with the pseudo labels generated in \ref{pseudo}, and a self-supervised refinement loss with Z-Stack image guidance which aligns the edges of the segmentation with cells borders. Our proposed loss reads as
\begin{equation}
    \mathcal{L}_{S2} = \mathcal{L}_{class} + \lambda_{BF}\mathcal{L}_{BF}
\end{equation}

\noindent where $\mathcal{L}_{class}$ is a Cross Entropy loss against our pseudo labels, $\lambda_{BF}$ is a weighting coefficient and $\mathcal{L}_{BF}$ is a BF refinement loss which we introduce to improve upon the pseudo labels.

\subsubsection{BF Refinement Loss}
Blob detection methods are frequently used in efficient image processing pipelines to detect regions with an appearance that differs from their surrounding \cite{yoo2019pseudoedgenet}.
The area information they capture is usually retrieved with a discretized version of the differentiable operator given by $\Delta = \nabla \cdot \nabla = \partial_x^2 +  \partial_y^2 +  \partial_z^2$.
Since the calculation of second order derivatives is very sensitive to noise, we robustify the calculation by applying it on a Gaussian-smoothed version of our input.\cite{lindeberg2015image}

\begin{figure}[htpb]
\centering
\includegraphics[width=\textwidth]{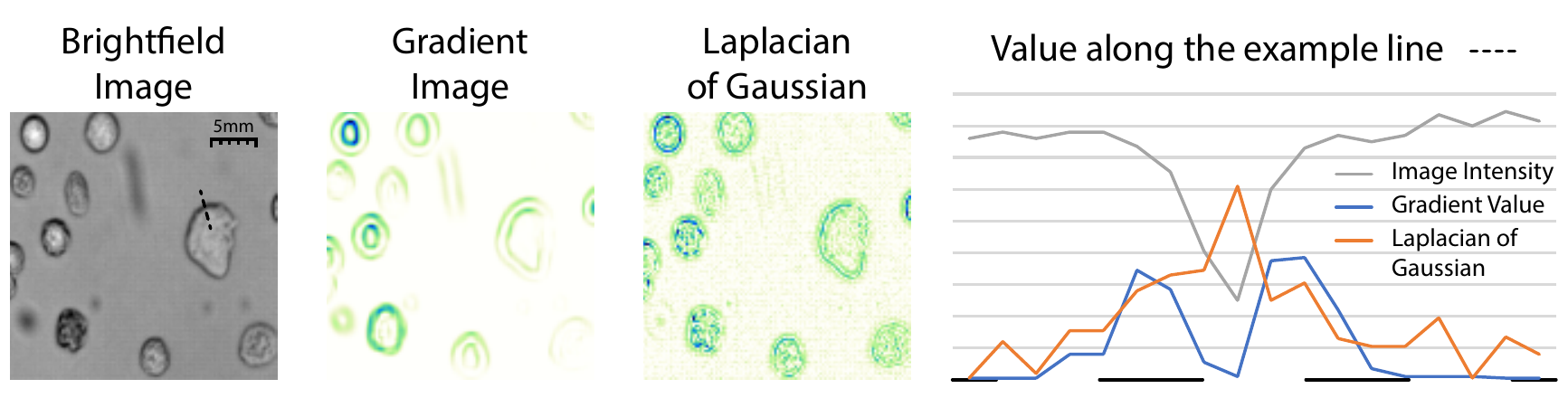}
\caption{\textbf{BF Boundary Loss.} An Z-Stack BFM image is shown (1st from left) together with its gradient image (2nd) and its LoG (3rd). The plot (right) indicates the value distribution along the dashed line for all three images. It can be seen that the LoG peaks at the cell boundary.}
\label{pics:boundary_loss}
\end{figure}

Inspired by the weighting mask for colour similarities based on image gradients proposed in \cite{heise2013pm}, we design a smooth loss term to align the segmentation mask with the visible cell membrane from the image using the LoG operator. Inspection of the image gradient along the cell boundary as depicted in Fig.~\ref{pics:boundary_loss} indicates that the segmentation boundary coincides with with the image Laplacian. To enforce accurate segmentation boundaries, we asymmetrically penalize a disagreement between the image $I$ and its segmentation mask $S$ by
\begin{equation}
    \mathcal{L}_{BF} = \nabla \text{G} \left(S, \sigma_1 \right) \cdot \exp \left(-\left\| \Delta \text{G} \left(I, \sigma_2 \right) \right\|_p^p\right),
    \label{bf_loss}
\end{equation}
where $\text{G}$ describes Gaussian smoothing with standard deviation $\sigma$ and $\nabla$ and $\Delta = \nabla \cdot \nabla$ are the first and second order differential operators.

\subsection{Instance Label Assignment}

With the cell centers likelihood map from \ref{pseudo} and the segmentation mask from $S2$, we first merge the peaks in the likelihood map with low Euclidean distance by a weighted mean over their scores and then run a watershed algorithm to assign instance labels.

%% file: 5-0-experiments.tex
Our models were trained using a GeForce GTX 1060 GPU with a learning rate of $5 \times 10^{-5}$ and weight decay of $10^{-6}$ using the Adam optimizer and a batch size of one due to the memory restrictions. We trained all models until convergence with a maximum of 40 epochs.

\subsection{BFM Z-Stack}
In absence of a publicly available dataset for this use case, we collected a BFM Z-Stack dataset of SF-9 insect cells (ThermoFisher, Germany) which we roughly annotated with geometric cell centers. In order to decrease the sedimentation speed of the cells, cells were suspended in feeding medium supplemented with 20\% Fetal Calf Serum (FCS). BFM images were obtained using a Zeiss Axiovert 40 CFL microscope with 20x magnification.


The dataset consists of 120 instances of $64\times256\times256$ with z-spacing of 2 $\mu m$ per slice. We crop the instances to 16 slices for training and resize each plane to a $128\times128$ image. The data is split to a 4:1 train/validation split. The dataset will be made publicly available upon acceptance.
\begin{figure}[htpb]
\centering
\includegraphics[width=\textwidth]{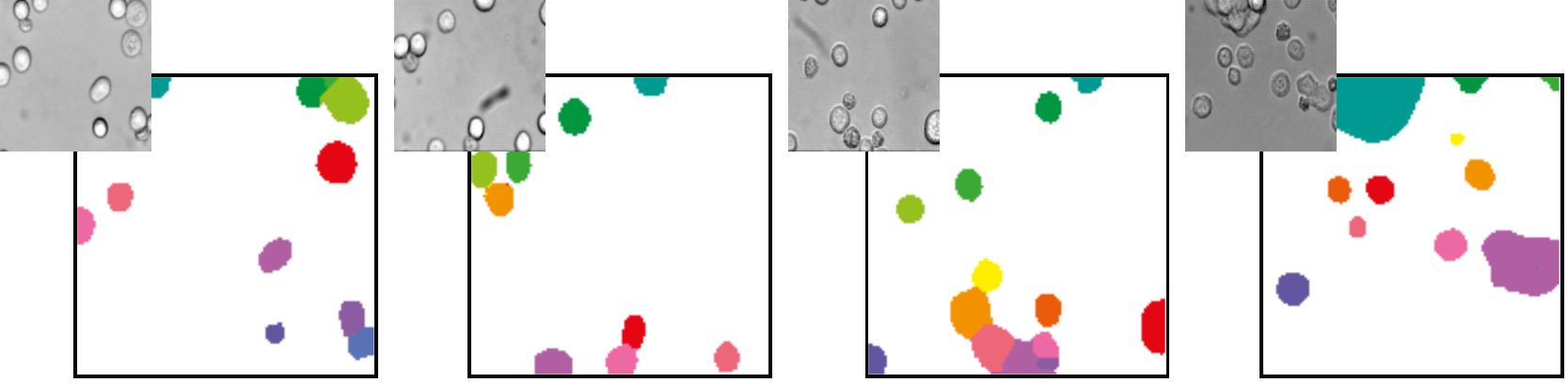}
\caption{\textbf{Example image from BFM Z-Stack.} Shown are four instance segmentation results with their corresponding BFM image.}
\label{pics:brightfield_results}
\end{figure}

We trained our model with the Euclidean norm $p=2$, $\lambda_{BF}=1$, and $\sigma_1 = \sigma_2 = 3$ in eq.\ref{bf_loss}. Without instance labels in this dataset, we evaluate our results on BFM Z-Stack qualitatively. (Fig.~\ref{pics:brightfield_results}).
While we see aligned segmentation and clear class separation on the first three results, the last indicates the limitation of our weakly supervised method in the presence of cell conglomerates where the segmentation mask provides cell regions for connecting cells whose instances labels become fused.

\subsection{Fluorescent Z-Stacks}

In the absence of another independent BFM Z-Stack dataset with cell annotations and to demonstrate the generalization capabilities of our approach, we used images from BBBC050~\cite{tokuoka20203d,ljosa2012annotated} which is a 3D image dataset of early mouse embryos with nuclei fluorescently labeled with mRFP1 and imaged with an IX71 microscope.

This allows to evaluate the results also quantitatively in comparison to the recent cell segmentation approach presented by~\cite{tokuoka20203d}. We generate the weak labels we need in our pipeline as average of each segmentation mask. Following their evaluation protocol, we report on the standard metrics for intersection over union (IoU) for segmentation as well as SEG~\cite{mavska2014benchmark} and MUCov~\cite{silberman2014instance} for instance labels. While SEG is an indicator of the absence of false-negative instances, MUCov indicates the absence of false-positives.

Since BBBC050 is a fluorescent microscopy dataset its appearance is different from BFM. Cell edges are not high response regions for LoG and the image statistics vary in comparison to BFM. To cope with this noticeable difference of the fluorescent images, we change our asymmetric boundary loss to penalize on gradient alignments instead of the Laplacian.
The loss then reads as:
\begin{equation}
    \mathcal{L}_{FL} = \nabla \text{G} \left(S, \sigma_1 \right) \cdot \exp \left(-\left\| \nabla \text{G} \left(I, \sigma_2 \right) \right\|_p^p\right).
\end{equation}
Again we choose $p = 2$, $\lambda_{FL}=1$, and $\sigma_1 = \sigma_2 = 3$. 

\begin{figure}[htpb]
\centering
\includegraphics[width=\textwidth]{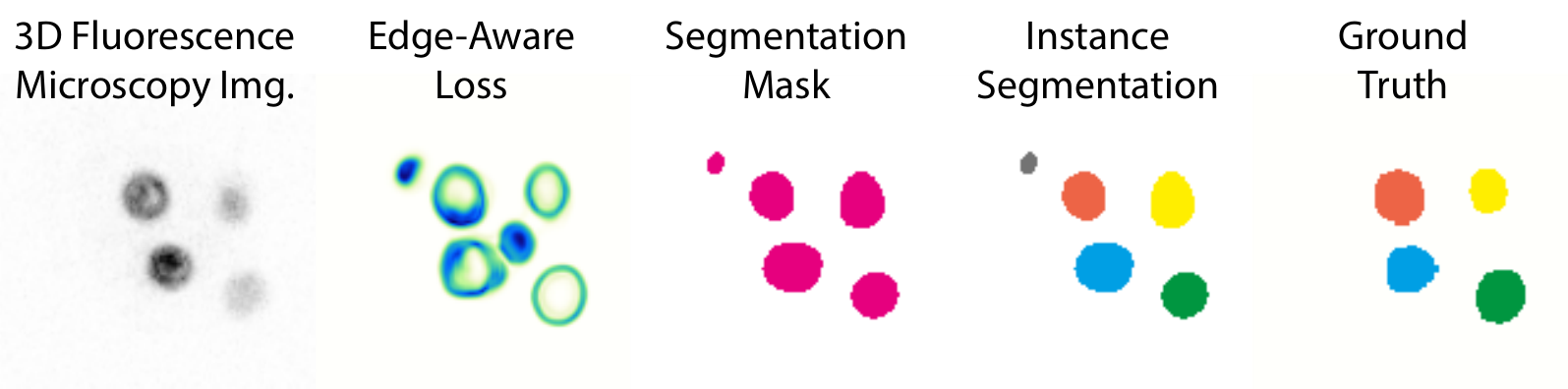}
\caption{\textbf{Example image from BBBC050 dataset.} An Image from the fluorescent Z-Stack is shown (1st from left) together with an evaluation of the Loss (2nd), the Segmentation Mask (3rd), our Instance Prediction (4th) and the Ground Truth (right).}
\label{pics:fluorescent}
\end{figure}

There is an overview of this loss and the areas it is effecting on a segmentation mask on an example image in Fig.~\ref{pics:fluorescent} together with our predictions.

We train our architecture with weak labels from the centroid of instances.
To assess the influence of the second network and the boundary loss, we ablate our full pipeline and compare it against the recent method QCANet~\cite{tokuoka20203d} and a 3D Mask R-CNN~\cite{he2017mask} as proposed in~\cite{tokuoka20203d}. (Results in Table.~\ref{tab:evaluation})

\begin{table}[htbp]
    \centering
    \caption{Comparison on BBBC050.}
    \begin{tabular*}{\textwidth}{l@{\extracolsep{\fill}}lllll}
        \toprule
		Method & Annotation Level & IoU & SEG & MUCov\\
		\midrule
		Fully supervised methods\\
		\midrule
        3D Mask R-CNN & Full Mask & 0.558 & 0.476 & 0.607\\
        QCANet~\cite{tokuoka20203d} & Full Mask & \textbf{0.746} & \textbf{0.710} & \textbf{0.721}\\
        \midrule
		Our training with weak annotations\\
		\midrule
        Our Pseudo Labels & Cell Centers & 0.506 & 0.353 & 0.399\\
        BFS-Net (+ $\mathcal{L}_{class}$) & Pseudo Labels & 0.603 & 0.441 & \textbf{0.506}\\
        BFS-Net full pipeline (+ $\mathcal{L}_{FL}$) & Pseudo Labels & \textbf{0.643} & \textbf{0.530} & 0.473\\
		\bottomrule
	\end{tabular*}
    \label{tab:evaluation}
\end{table}

While the performance of our network trained with weak labels results in a $0.1$ lower IoU compared to the fully supervised method QCANet, we are able to outperform the fully supervised 2-stage pipeline of 3D Mask R-CNN in terms of IoU as reported by~\cite{tokuoka20203d}.
The result of training with weak labels reaches $86.2\%$ IoU performance in comparison to the fully supervised counterpart 
QCANet while requiring significantly fewer annotation effort. However shows a larger gap on the instance metrics.
We believe that this is due to the refinement loss refining the segmentation quality while not specifically targeting the instance labels.
As expected, the performance of using pseudo labels leads to the lowest performance. If we train $S2$ without BF-Refinement Loss, the performance increases. We believe that this is due to the training helping to remove false positives. When $S2$ is trained with BF-Refinement Loss, the result gets improves further in terms of segmentation alignment due to the Z-Stack guided refinements.

%% file: 6-0-discussion.tex
We have proposed the first weakly supervised approach for cell instance segmentation of BFM Z-Stack together with a new dataset.
Rough center points annotations were used to create pseudo labels which we refined using a novel loss.
We demonstrated the qualitative and quantitative performance of our method on two datasets where we reach an IoU score of $86.2\%$ compared to the state-of-the-art method QCANet despite significantly fewer label effort.
We believe that our dataset can pave the way to more methods based on weak labels in the domain of Bright-Field Microscopy.

